\documentclass[10pt,twocolumn,letterpaper]{article}

\usepackage{wacv}
\usepackage{times}
\usepackage{epsfig}
\usepackage{graphicx}
\usepackage{amsmath}
\usepackage{amssymb}
\usepackage{enumitem}
\usepackage{algorithm}
\usepackage{algpseudocode}
\usepackage{array}
\usepackage{subfigure}
\usepackage{epstopdf}
\usepackage[final]{changes}

\newcommand{\bbR}{\mathbb{R}}

\newcommand{\cnet}{\emph{ClusterNet }}
\newcommand{\calC}{\mathcal{C}}
\newcommand{\calX}{\mathcal{X}}
\newcommand{\calL}{\mathcal{L}}
\newcommand{\calT}{\mathcal{T}}

\newcommand{\bC}{\mbox{$\mathbf{C}$}}

\wacvfinalcopy

\usepackage[breaklinks=true,bookmarks=false]{hyperref}
\definechangesauthor[name=Saket Anand, color=blue]{SA}

\def\wacvPaperID{196} 

\ifwacvfinal\fi
\begin{document}

\title{Semi-Supervised Clustering with Neural Networks}

\author{Ankita Shukla \hspace{2cm} Gullal Singh Cheema \hspace{2cm} Saket Anand \\
IIIT-Delhi, India \\
{\tt\small \{ankitas, gullal1408, anands\}@iiitd.ac.in}
}

\maketitle
\ifwacvfinal\thispagestyle{empty}\fi

\begin{abstract}
 Clustering using neural networks has recently demonstrated promising performance in machine learning and computer vision applications. However, the performance of current approaches is limited either by unsupervised learning or their dependence on large set of labeled data samples.
In this paper, we propose \cnet that uses pairwise semantic constraints from very few labeled data samples ($<5\%$ of total data) and exploits the abundant unlabeled data to drive the clustering approach. 
We define a new loss function that uses pairwise semantic similarity between objects combined with constrained $k$-means clustering to efficiently utilize both labeled and unlabeled data in the same framework. The proposed network uses convolution autoencoder to learn a latent representation that groups data into $k$ specified clusters, while also learning the cluster centers simultaneously. 
We evaluate and compare the performance of \cnet on several datasets and state of the art deep clustering approaches. 
\end{abstract}

\section{Introduction}
Clustering is a classical unsupervised learning problem that seeks insights into the underlying structure of data by naturally grouping similar objects together. The notion of similarity often depends on the distance measure defined in some feature representation space, and is a crucial factor that governs the performance of a clustering algorithm. Consequently, learning both, a meaningful representation space as well as an appropriate distance measure from the data \cite{Xing_NIPS2003,Bilenko_ICML2004} has shown to significantly improve clustering performance. While both these factors are centric to clustering, they play an equally important role in supervised tasks like classification and detection. As a result, on one hand, there have been recent advances in supervised representation learning
\cite{Krizhevsky_ImageNet2012,Malik_ECCV2014,Girshick_CVPR2014} where deep neural networks (DNNs) have achieved state of the art performance in classification, detection and semantic segmentation problems. On the other hand, DNNs have also proved effective in unsupervised learning with architectures like autoencoders, which learn representations that consistently reconstruct the input.
This capability has led to deep clustering methods\cite{Xie_ICML2016,Dizaji_ICCV2017,Guo_IJCAI2017,Yang_Kmeans_space2016} that rely on autoencoders, or on convolutional neural networks (CNNs) \cite{Kira_2015,Batra_CVPR2016}. For applying DNN based models to clustering, two approaches are commonly used: methods that explicitly compute cluster centers \cite{Xie_ICML2016,Guo_IJCAI2017,Yang_Kmeans_space2016} and methods that directly or indirectly model the data distribution \cite{Kira_2015,Dizaji_ICCV2017}.

\begin{figure}[h!]
	\begin{center}
		\includegraphics[trim=2cm 1cm 2cm 1cm,clip=true,width =0.15\textwidth]{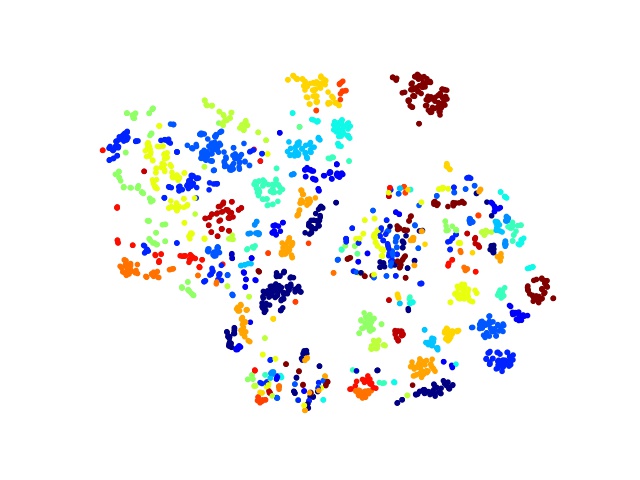}
        \includegraphics[trim=2cm 1cm 2cm 1cm,clip=true,width =0.15\textwidth]{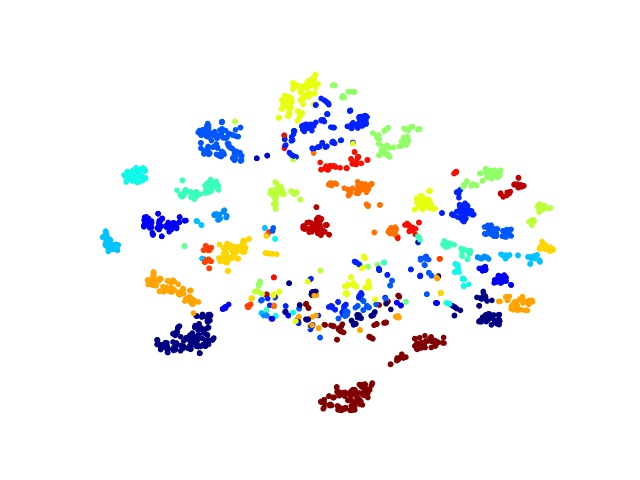}
        \includegraphics[trim=2cm 1cm 2cm 1cm,clip=true,width =0.15\textwidth]{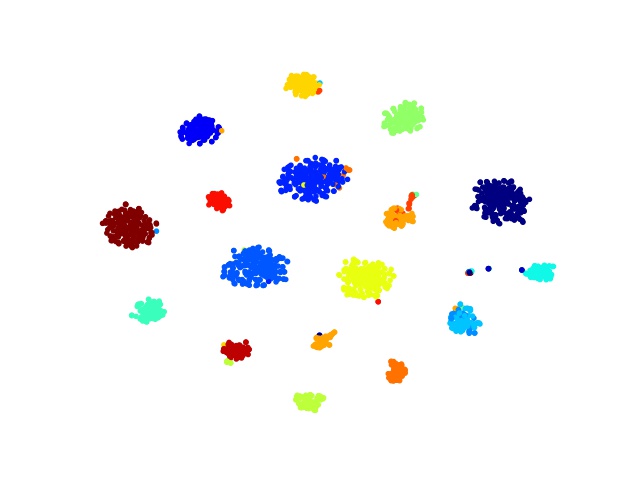}
		\caption{T-sne plots for FRGC dataset: Raw data (Left Image), after unsupervised training of autoencoder (Center Image) and after training with clustering loss (Right Image, 2\% labeled data)}
		\label{fig:before_after}
		\end{center}
\end{figure}

While clustering has traditionally been an unsupervised learning problem, many DNN based clustering approaches have incorporated varying degrees of supervision, which can substantially improve the clustering performance. Based on the level of supervision, clustering approaches can be categorized as unsupervised, supervised and semi-supervised. Fully supervised approaches \cite{Law_ICML2017,Song_CVPR2017} perform well but are often limiting due to unavailability of sufficient labeled data. On the other hand, approaches like \cite{Xie_ICML2016,Guo_IJCAI2017} are unsupervised and rely on unlabeled data alone for learning the network parameters, assuming only the knowledge of the number of clusters. Fully unsupervised approaches, while very desirable, are oblivious of any semantic notions of data, and may lead to learned representations that are difficult to interpret and analyze.
Weakly or semi-supervised approaches are often used as a reasonable middle-ground between the two extremes. For example, \cite{Kira_2015} uses pairwise constraints as a weaker form of supervision, and \cite{Fogel_pair2018} exploit both, labeled as well as unlabeled data for learning.  






In this work, we introduce \emph{ClusterNet}, an autoencoder based architecture that learns latent representations suitable for clustering. The training works in a semi-supervised setting utilizing the abundant unlabeled data augmented with pairwise constraints generated from a few labeled samples. An example is shown in Fig \ref{fig:before_after}, that shows the effect of using small amount of labeled data on the feature embedding space. In addition to the usual reconstruction error, ClusterNet uses an objective function that comprises two complementary terms: a $ k $-means style clustering term that penalizes high intra-cluster variance and a pairwise KL-divergence based term that encourages similar pairs to have similar cluster membership probabilities.


Our contributions are summarized below:
\begin{itemize}
\item ClusterNet is an semi-supervised clustering approach that relies on a simple convolutional autoencoder architecture to learn \emph{clusterable} latent representations in an end-to-end fashion. 
\item ClusterNet uses a loss function that combines a pairwise divergence defined over assignment probabilities with a $ k $-means loss that complement each other to simultaneously learn the cluster centers as well as semantically meaningful feature representations.
\item ClusterNet requires minimal hyperparameter tuning and is easy to train, due to an annealing strategy that adaptively trades off the importance of labeled and unlabeled data during the training. 

\end{itemize}

The remainder of the paper is organized as follows. We discuss the related work in Section \ref{sec:rel_work}. In Section \ref{sec:bg}, we discuss the preliminary ideas underlying our work followed by the details of our proposed approach in Section \ref{sec:prop_work}. We present the experimental evaluation in Section \ref{sec:expts} and finally conclude in Section \ref{sec:conc}.

\section{Related Work}\label{sec:rel_work}
In this section, we restrict our discussion to recent progress in clustering in a deep learning framework. Many of the clustering approaches leverage the autoencoder due to its ability to learn representations in an unsupervised manner.
Xie \emph{et al.} \cite{Xie_ICML2016} employed the encoder of a layer-wise pretrained denoising autoencoder to obtain both the latent space representations and cluster assignment simultaneously. Guo \emph{et al.} \cite{Guo_IJCAI2017} improved upon this approach by incorporating an autoencoder to preserve the local structure of the data.
While these two approaches use fully connected multi-layer perceptron (MLP) for encoder and decoder architectures, work in \cite{Gao_CNN2017} use a convolutional autoencoder to effectively handle image data. Apart from encoder-decoder architectures, convolutional neural network (CNN) based architectures have also been applied to the clustering problem. For example, the work in \cite{Batra_CVPR2016} proposed JULE takes an agglomerative clustering approach that utilizes a CNN based architecture. \added[id=SA,remark={what kind of hyperparameters. Do they mention it? \textcolor{red}{6 hyperparameters, table 2 in their paper}} ]{While this approach performs well it requires a large number of hyper-parameter tuning, limiting its applicability in real world clustering problem.} Further, being fully unsupervised, these approaches may not learn a semantically meaningful representation, which can subsequently affect the clustering performance. Huang \emph{et al.} \cite{Dizaji_ICCV2017} proposed an unsupervised approach, DEPICT, that also used a convolutional autoencoder stacked with a softmax layer in the latent space. A KL-divergence loss was used to optimize for the cluster assignments in addition to a regularization term for discovering balanced clusters. \added[id=SA,remark={why do we make this claim?; it may need more explanation \textcolor{red}{it forces balanced distribution of the samples by adding a term in the objective function}}]{Such a loss tends to limit its capability to handle uneven class distributions.} While this approach extends to the semi-supervised learning scenario and shows out of sample generalization, it requires fine-tuning the network with cross entropy loss as done in \cite{Rasmus_NIPS2015,Zhao_2015}.




More recently, several deep learning approaches \cite{Rasmus_NIPS2015,Zhao_2015,Kingma_NIPS2014} have utilized both labeled and unlabeled data for classification, however semi-supervised clustering is not well explored. Work in \cite{Kira_2015} uses pairwise constraints on the data and enforces small divergence between similar pairs while increasing the divergence between dissimilar pair assignment probability distributions. However, this approach did not leverage the unlabeled data, leading to suboptimal performance in small labeled data settings. Fogel \emph{et al.} \cite{Fogel_pair2018} proposed a clustering approach using pairwise constraints, that are obtained by considering a mutual KNN graph on the unlabeled data for unsupervised clustering. The method further extends to semi supervised approach by using few labeled connections that are defined using the distances defined on the embedding obtained from initial autoencoder training. However, several parameters like distance threshold and number of neighbors required for MKNN are user defined and are crucial to the performance of the algorithm.



\section{Background}\label{sec:bg}
We leverage the k-means clustering framework with KL divergence based soft pairwise constraints. We briefly discuss these ideas here and develop our proposed semi-supervised clustering framework in the following section. 

\subsection{Constrained KMeans}

\cnet is inspired by the constrained $ k $-means algorithm \cite{Basu_2002} (Algo. \ref{algo: ckmeans}) that utilizes some labeled data to guide the unsupervised $ k $-means clustering. As opposed to random initializations of cluster centers in traditional $ k $-means, labeled samples are used to initialize the cluster centers in constrained $ k $-means. Secondly, at each iteration of $ k $-means the cluster re-assignment is restricted to the unlabeled samples, while the membership of labeled samples is fixed. This procedure of constrained $ k $-means has shown performance improvement over $ k $-means algorithm. 

\begin{algorithm}[h!]
	{\small
		\caption{Constrained KMeans \cite{Basu_2002}}
		\label{algo: ckmeans}
		\textbf{Input: }  $\calX^u$ = $\{x^{u}_1, x^{u}_2, \cdots x^{u}_m\}$: set of unlabeled data points, $\calX_{k}^{l}$ = $ \{x^{l}_1, x^{l}_2, \cdots x^{l}_p\}$: Set of labeled data in class k, $\calX^l$ = $\cup_{k=1}^{K} \calX_k^l$\ \\
 		\textbf{Output:} Disjoint $K$ sets $\{\calC_k\}_{k=1}^{K}$ of $\calX^u$ that minimizes the KMeans objective function\\
        \textbf{Method:}\vspace{-0.2cm}
        \begin{enumerate}
        \item $t = 0$ \vspace{-0.15cm}
        \item Initialize cluster centers
        $\mu_k = \frac{1}{|\calX_k^l|}\sum_{x \in \calX_k^l }{x}$ \vspace{-0.2cm}
        \item Repeat till convergence \vspace{-0.2cm}
        	\begin{itemize}
            \item Assign data to cluster\\
            For labeled data : \\
            $x \in \calX_k^l$ assign $x$ to cluster $\calC_k^{t+1}$ cluster \\
            For unlabeled data:\\ 
            for $x_i^u \in \calX^u$ assign to $\calC_k^{t+1}$ cluster obtained by 
            
            $k = \arg \min_{k} ||x_i^u-\mu_k^{t}||^2 $ \vspace{-0.15cm}
            \item Update centers :
            $\mu_k^{t+1}= \frac{1}{|\calC_k^{t}|}\sum_{x \in \calC_k^{t} }{x}$ \vspace{-0.2 cm}
            \item $t \leftarrow t+1$
            \end{itemize}
        \end{enumerate}
}
\end{algorithm}

\subsection{Pairwise KL Divergence}
Another component of \cnet is based on the KL divergence, which is a measure of disparity between two probability distributions. The KL divergence has often been used in clustering approaches \cite{Xie_ICML2016,Guo_IJCAI2017,Dizaji_ICCV2017,Kira_2015}. Given $ K $-dimensional vector of cluster assignment probabilities $ p $ and $ q $ corresponding to points $x_p$ and $x_q$ respectively, the KL divergence between $p$ and $q$ is given by 
\begin{align}
KL(p||q) =\sum_{i=1}^{K}{p_i \log\frac{p_i}{q_i}}
\end{align}
Here, $K$ is the number of clusters\deleted[id=SA]{, or the number of output nodes in the softmax layer}. Similar to Kira \emph{et al.} \cite{Kira_2015}, we use a \added[id=SA,remark={Is this our contribution, or people have used it? \textcolor{red}{cited Kira}}]{symmetric variant} of the KL divergence since we are only concerned with optimizing the loss function for $p$ and $q$ simultaneously
\begin{align}
\calL_{p,q} =KL(p||q)+KL(q||p)
\end{align}
The loss is obtained by first \added[id=SA,remark={what does it mean to fix $p$? Is $ q $ varying when $ p $ is fixed?}]{fixing} $p$ and computing the divergence of $q$ from $p$ and vice versa.

  


\section{Proposed Approach}\label{sec:prop_work}

In this section, we present our approach: the network architecture, loss function and explain the optimization strategy. We consider the following setup for \cnet. Let $\calX^l =\cup_{k=1}^{K}\{\calX_k^l\}$ denote the set of labeled data corresponding to $K$ clusters. \added[id=SA,remark={Should there be a superscript here $ x_1^l $? \textcolor{red}{Most equations don't have it, will add it}}]{Here $\calX_k^l =\{x_1^l,x_2^l..x_p^l\} \in \bbR^n$ is the set of labeled points in $k^{th}$ class}. Let $\calX^u$ denote a set of unlabeled data points. The goal is to learn a feature representation that is amenable to forming $K$ clusters in a manner that similar pairs of points tend to belong to the same cluster while the dissimilar pairs do not.
\subsection{Network Architecture}
We build our clustering model using a convolutional autoencoder architecture.
The parameters for encoder and decoder are represented by $\theta_e$ and $\theta_d $ respectively.
The cluster centers are given by $\mu \in \bbR^d$, where $d$ is the dimensionality of the latent space. We denote the cluster membership of the $i^{th}$ point using a one-hot encoding vector $a_i\in\{0,1\}^K$, such that $a_{i,j}=1$ implies that the $i^{th}$ point lies in the $j^{th}$ cluster. Further, we use $z=f(\theta_e; x) $ to denote latent space representations, where $f(\theta_e; \dot): X \rightarrow Z$ is a nonlinear mapping from the input space to the latent space. Lastly, $g(\theta_d, z)$ represents the output of the autoencoder that maps the latent space representations to the reconstructed input space. For all our experiments, we have fixed both the network architecture as well as the dimension of the latent space to show that clustering performance of \cnet is sustained across multiple datasets.

\subsection{Loss Function}\label{sec:obj_fn}
Our semi-supervised clustering loss function has three components: pairwise loss, cluster loss and a reconstruction loss, each defined for both labeled as well as unlabeled data and is defined as 
\begin{align}\label{eq:loss}
\calL =  \calL_{pair}^{l} + \calL_{cluster}^{l} + \lambda(\calL_{pair}^{u} + \calL_{cluster}^{u} ) +\calL_{recon}
\end{align}
The reconstruction loss is common for all data, but the cluster and pairwise losses are defined slightly differently for the labeled and the unlabeled data, denoted by the superscripts $ l $ and $ u $ respectively in (\ref{eq:loss}). The reconstruction loss serves as a regularizer for ensuring the representation has high fidelity, while the other two terms make the latent space more amenable to forming semantically consistent clusters. Here, $\lambda$ acts as a balancing coefficient and is important for the performance of our algorithm. A high value of $\lambda$ suppresses the benefits of labeled data, whereas a very small value of $\lambda$ fails to use the unlabeled data effectively during training. Therefore, we resort to a deterministic annealing strategy \cite{Grandvalet_2006, Lee_2013}, that gradually increases the value of $\lambda$ with time and tends to avoid poor local minima. Next, we discuss each component of the loss term in detail.

\noindent \textit{\textbf{Cluster Loss:}} Similar to $ k $-means loss term, the cluster loss aims to minimize the distance between latent space representations of samples with the corresponding cluster centers to encourage clusterable representations. The corresponding loss for both labeled and unlabeled data are given by
\begin{align}
\calL_{cluster}^u = \frac{1}{|\calX^u|}\sum_{x \in \calX^u}{||f(\theta_e;x_i^u)-\bC a_i^u||_2^2}
\end{align}
Here, $\bC \in \bbR^{d \times K}$ is a matrix with each column corresponding to one cluster center. Similarly, for the labeled data $\calX^l$
\begin{align}
\calL_{cluster}^l = \frac{1}{|\calX^l|}\sum_{x \in \calX^l}{||f(\theta_e;x_i^l)-\bC a_i^l
||_2^2}
 \end{align}
Here, $a^u$ is the \emph{predicted} label assignments for unlabeled data and $a^l$ is the \emph{true} label for labeled data.\\
\noindent \textit{\textbf{Pairwise Loss:}}
The KL-divergence loss utilizes similar and dissimilar pairs and encourages similar cluster assignment probabilities for similar point pairs, while ensuring a large divergence between assignment probabilities of dissimilar pairs.

For labeled data, the pairwise loss is given by 
\begin{align}\label{eqn:pair_label}
\noindent\calL_{pair}^l \!&\!=\! \frac{1}{|\calT_{sim}^l|}\hspace{-0.1cm}\sum_{(p,q)\in \calT_{sim}^l}\hspace{-.35cm}\calL_{sim} \!+ \!\frac{1}{|\calT_{dsim}^l|}\underset{{(p,q)\in \calT_{dsim}^l}}\sum\hspace{-0.35cm}\calL_{dsim}\\
\nonumber \text{where}\\ 
\nonumber \calL_{sim} &= KL(p||q)+KL(q||p)\\
\nonumber \calL_{dsim} &= [0,m-KL(p||q))]_+ + [0,m-KL(q||p)]_+
\end{align}
Here, $p$ and $q$ are $ K $-vectors denoting the cluster assignment probabilities for points $f(\theta_e;x_p)$ and $f(\theta_e;x_q)$ obtained based on the distances from the cluster centers. The margin $m$ is introduced to impose a minimum separability between dissimilar pairs. The sets, $\calT_{sim}^l$ and  $\calT_{dsim}^l$ are the similar and dissimilar image pairs respectively generated using the labeled data. Similarly, the labels based on cluster membership are used to compute the pairwise loss for unlabeled data as 
\begin{align}\label{eqn:pair_unlabel}
\noindent\calL_{pair}^u \!&\!= \!\frac{1}{|\calT_{sim}^u|}\hspace{-0.1cm}\sum_{(p,q)\in \calT_{sim}^u}\hspace{-0.35cm}\calL_{sim} \!+\! \frac{1}{|\calT_{dsim}^u|}\hspace{-0.1cm}\underset{{(p,q)\in \calT_{dsim}^u}}\sum\hspace{-0.35cm}\calL_{dsim}
\end{align}
Here $\calL_{sim}$ and $\calL_{dsim}$ are defined as before.

\noindent \textit{\textbf{Reconstruction Loss:}}
While we want the latent space representations to form clusters, the cluster and pairwise objective can lead to a degenerate latent space where points tend to collapse to the cluster center, in an attempt to minimizing the cluster loss. This degeneracy may lead to poor generalization to unseen points. To mitigate this effect, we regularize the clustering loss by adding a reconstruction loss term for the autoencoder given by 
\begin{align}
\calL_{recon} = \sum_{x \in \{\calX^u \cup \calX^l\}}{||g(f(x_i))-x_i||^2_2}
\end{align}

\subsection{Network Optimization}
We alternately optimize the network parameters and the cluster center parameters. For fixed cluster center parameters \emph{i.e.} $\bC$ and $a$, we update the network parameters $\theta_e$ and $\theta_d$ using standard backpropagation algorithm. The steps of our algorithm are summarized in Algo. \ref{algo:clusternet}.
\subsubsection{Cluster Center Parameters}
\noindent \textit{\textbf{Cluster Center Initialization:}}
Motivated by constrained $k$-means, we initialize the cluster centers by mean of the labeled samples in the corresponding labeled set $\calX^l$
\begin{align}
\mu_k = \frac{1}{|\calX_k^l|}\sum_{|x \in \calX_k^l|}{f(\theta_e; x_i^l)} \quad for  \; k = 1,\cdots K 
\end{align}

\noindent \textit{\textbf{Cluster Assignment:}}
For given cluster centers $\bC = \{\mu_1, \mu_2 \cdots \mu_K \} $ and non linear embeddings in the latent space, similar to constrained $k$-means the class or cluster associated with a data sample $x_i$ is obtained by eqs. (\ref{eq: asgn_unlab}) and (\ref{eq: asgn_lab})  for unlabeled and labeled data respectively.
\begin{equation}
  \qquad a_{j,i}^u = \begin{cases}
1 \quad\text{if}  \; j = \hspace{-0.2cm}\underset{k =\{1, 2, \cdots K\}}{\arg\min}\! ||f(\theta_e;x_i^u)\!-\!\mu_k||\\
   0 \quad  \text{ otherwise } 
  \end{cases}
  \label{eq: asgn_unlab}
\end{equation}
\begin{equation}
  \qquad a_{j,i}^l = \begin{cases}
1 \quad\text{for} \; j = k \quad \text{for} \quad x_i^l \in \calX_k^l \\
   0 \quad  \text{ otherwise } 
  \end{cases}
  \label{eq: asgn_lab}
\end{equation}

\noindent \textit{\textbf{Cluster Center Updates:}}
A naive way to update cluster centers is to follow the $ k $-means strategy and compute the mean of the samples allocated to a cluster by looking at the current values of $a^u$ and $a^l$. This approach can pose a serious challenge due to  a dynamic representation space. While $a_i^l$ is informative and denotes the true cluster membership, the predicted unlabeled data assignments $ a_i^u $ may not be accurate and can introduce a significant bias in the updated centers.

Therefore, similar to \cite{Yang_Kmeans_space2016}, we update cluster centers using gradient updates with an adaptive learning rate.
Further, we utilize the labeled data to update the corresponding true cluster center as 
 \begin{align}
\mu_k \leftarrow \mu_k -\frac{1}{N_{k}^l}{(\mu_k-f(\theta_e, x_i^l))}\\
\nonumber \text{for } x_i \in \calX_k^l
\end{align}
The unlabeled data update the cluster centers corresponding to the predicted cluster label in eq. (\ref{eq: asgn_unlab})
\begin{align}
\mu_k \leftarrow \mu_k -\frac{1}{N_{k}^u}{(\mu_k-f(\theta_e, x_i^u))a_i^u}
\end{align}
Here, $N_k^l$ and $N_k^u$ are the number of labeled and unlabeled samples respectively, that have been assigned to cluster $k$ in the current iteration. 

\noindent \textit{\textbf{Pairwise Constraints}}:
The pairwise loss in eqs. (\ref{eqn:pair_label}) and (\ref{eqn:pair_unlabel}) is defined on similar and dissimilar image pairs. For labeled data, these pairs are created using the label information as follows 
\begin{align}
\nonumber &\calT_{sim}^l =\{(i, j): x_i \in \calX_{k_1}^l,x_j \in \calX_{k_2}^l, k_1 = k_2\},\\
\nonumber &\calT_{dsim}^l =\{(i, j): x_i \in \calX_{k_1}^l, x_j \in \calX_{k_2}^l, k_1 \neq k_2\},\\
&\qquad ~k_1,~k_2 \in \{1,2,\cdots,K\}
\end{align}
Similarly, the pairs are also computed for unlabeled data based on the predictions given by eq. (\ref{eq: asgn_unlab})
\begin{align}
\nonumber &\calT_{sim}^u =\{(i, j): x_i \in \calX_{k_1}^l, x_j \in \calX_{k_2}^l, k_1 = k_2\},\\
\nonumber &\calT_{dsim}^u =\{(i, j): x_i \in \calX_{k_1}^u, x_j \in \calX_{k_2}^l, k_1 \neq k_2\}, \\
&\qquad ~k_1,~k_2 = \{1,2, \cdots K\}
\end{align}

\noindent \textit{\textbf{Assignment Probabilities:}} In many approaches, for a given sample the cluster assignment probabilities are obtained by a softmax layer with learnable parameters \cite{Kira_2015,Dizaji_ICCV2017}. Instead, we simply define probabilities based on distances from the cluster centers as
\begin{align}
p_{k, i} &= \frac{\exp(-d_{k,i})}{\sum_{k=1}^{K}{\exp(-d_{k,i})}}; \; k =1,2, \cdots K\\
\nonumber \text{where} \quad  d_{k,i} &=||f(\theta_e;x_i)-\mu_k)||_2^2
\end{align}\label{eq :prob_value}
Here, $p_{k,i}$ is the probability of assigning the $i^{th}$ sample to the $k^{th}$cluster. 

\begin{algorithm}[t]
	{\small
		\caption{ClusterNet Optimization}
		\label{algo:clusternet}
		\textbf{Input:}  $\calX^u$= $\{x^{u}_1, x^{u}_2, \cdots x^{u}_m\}$ :set of unlabeled data points, $\calX_{k}^{l}$ = $ \{x^{l}_1, x^{l}_2, \cdots x^{l}_p\}$: Set of labeled data in class k, $\calX^l$ = $\cup_{k=1}^{K} \calX_k^l$\ \\
 		\textbf{Output:} $\bC = \{ \mu_1, \mu_2, \cdots \mu_k \}$: Cluster centers, $\theta_e$: encoder parameters\\
        \textbf{Method:}
        \begin{enumerate}
 \vspace{-0.15cm}
         \item Initialize cluster centers
        $\mu_k = \frac{1}{|\calX_k^l|}\sum_{x \in \calX_k^l }{x}$ \vspace{-0.2cm}
        \item for $t =1 \cdots T$ \vspace{-0.2cm}
        	\begin{enumerate} [label=\Roman*]
            \item Create Similar and Dissimilar pairs $\calT_{sim}^l$ and $\calT_{dsim}^l$ respectively from labeled data $\calX^l$
            \item Compute labels for unlabeled data $\calX^u$ using cluster assignments from eq. \ref{eq: asgn_unlab}
            \item Create $\calT_{sim}^u$ and $T_{dsim}^u$ using (II)
            
          \item Compute assignment probabilities for both labeled and the unlabeled data using eq. \ref{eq :prob_value}
          \item Compute cluster center updates with labeled and unlabeled data using eq.
          \item Update network parameters $\theta_e$ and $\theta_d$
          
            \end{enumerate}
	\item Return cluster centers $\bC={\mu_1,\mu_2, \cdots, \mu_K}$ and $\theta_e$
        \end{enumerate}
}
\end{algorithm}

\section{Experiments}\label{sec:expts}
In this section, we evaluate the performance of \cnet on several benchmark datasets and compare it with state of the art approaches. We present evaluation results of our approach for both clustering and classification problem.
\noindent \textbf{Datasets}
We use 5 datasets for experiments: two handwritten digits dataset; MNIST \cite{MNIST_1998} and USPS \footnote{\url{https://cs.nyu.edu/~roweis/data.html}} and three face datasets; FRGC \footnote{\url{https://sites.google.com/a/nd.edu/public-cvrl/data-sets}}, CMU-PIE \cite{CMU2002cmu} and YouTube Faces \cite{YouTube_2011}. The details of the datasets are given in the Table \ref{tab: datasets}. For YTF dataset, we use the first 41 subjects, sorted according to their names in alphabetical order as in DEPICT \cite{Dizaji_ICCV2017}. For FRGC dataset, as in \cite{Dizaji_ICCV2017} we use 20 randomly selected subjects for our experiments.

\begin{table*}[h]
{\small 
	\centering
		\begin{tabular}{|ccccccc|}
		\hline
         & MNIST-train &  MNIST-test& USPS & FRGC & CMU-PIE & YouTube Faces \\ \hline
 \# Samples &60.000&10,000& 11,000&2462&2856&10,000 \\
 \# Classes &10 & 10& 10& 20& 68& 41 \\
 Size  &$32 \times 32$ & $32 \times 32$ & $16 \times 16$ & $32 \times 32 \times 3$ &$32 \times 32 \times 3$ &$55 \times 55 \times 3$ \\\hline
        
        	\end{tabular}
            \vspace{0.1cm}
	\caption{Dataset Description}
	\label{tab: datasets}
    }
 \end{table*}
\begin{table*}[h]
{\small 
	\centering
		\begin{tabular}{|cccccc|}
		\hline
        &   MNIST-test& USPS & FRGC & CMU-PIE & YouTube \\ \hline
 \#Samples & 10,000 & 1100 & 246 & 285 & 1000 \\\hline
        
        	\end{tabular}
            \vspace{0.1 cm}
	\caption{Samples to evaluate the performance of \cnet on unseen data samples. These samples are new to the network and have not been used by the network for training the network for clustering or in the pretraining stage. }
	\label{tab: datasets_oose}
    }
 \end{table*}

\newcolumntype{P}[1]{>{\centering\arraybackslash}p{#1}}
		
  
 

\begin{table*}[h!]
{\small 
	\centering
		\begin{tabular}{|c|c|P{2cm}cP{2cm}cccc|}
		\hline		
 Datasets & Evaluation& DEC & JULE-SF& JULE-RC& DEPICT & CPAC-VGG& \multicolumn{2}{c|}{ClusterNet}  \\
  & & & & & & &  Avg & Best \\\hline
 MNIST-full		 & NMI & 0.816	&0.906	&0.913 &0.917&- & \textbf{0.926}& \textbf{0.941}\\
       		 	 & ACC & 84.4	&95.90	&96.4  &96.50	& - & \textbf{96.84 } & \textbf{97.83}\\
Labeled Data 	 & 	& -  &  -     & -     & -    & -   &   \textbf{0.5\%}  &\textbf{0.5\%}\\\hline\hline
      
 USPS     		& NMI & 0.586 & 0.858 &0.913 & 0.927 &0.914 & {0.923}& \textbf{0.930}\\
      			& ACC &61.9  & 92.2  & 95.0 & 96.4  &86.7&   \textbf{96.51}& \textbf{96.9}\\
 Labeled Data 	&&  -   & -     & -     & -    &   5k connect     &  \textbf{ 2 \%}&\textbf{ 2 \%} \\     \hline\hline

 FRGC     		& NMI & 0.504 & 0.566  &0. 574 & 0.610 & 0.799&\textbf{0.850}& \textbf{0.947}\\
      			& ACC &	37.8  & 46.1   & 46.1  &  47.0 & 54.0& \textbf{81.16} & \textbf{91.92}\\
 Labeled Data 	& 	& - &   -   &    -   & -     &    5k connect   & \textbf{2 \%}& \textbf{2 \%}\\     \hline\hline

 CMU-PIE     	& NMI & 0.924 & 0.964 & \textbf{1.00} & 0.974 & 0.849&\textbf{1.00} &\textbf{1.00} \\
      			& ACC& 80.1  & 98.0  & \textbf{100} & 88.3 & 68.8 &\textbf{1.00} &\textbf{100}	\\
 Labeled Data 	& 	&-  & -     & -     &  -   &  5k connect    &  \textbf{2\%}& \textbf{2\%}\\      \hline\hline
 
 YouTube Face  	& NMI &0.446 & 0.848 & 0.848 & 0.802 &0.860& \textbf{0.932} & 
 \textbf{0.948}\\
      			& ACC &	37.1  & 68.4  &  68.4 & 62.1 &54.2 & \textbf{90.31} & \textbf{93.15} \\
 Labeled Data 	&     &-& -     & -     & -     &  5k connect    & \textbf{2 \%}  & \textbf{2 \%}\\     \hline
    \end{tabular}
    \vspace{0.1cm}
	\caption{Performance comparison of different algorithms on several datasets based on NMI (normalized mutual score) and ACC (Accuracy in \%). The results for various approaches are reported from the original paper. Compared unsupervised approaches use no labeled data, we put a dash mark (-) to indication. The semi supervised approaches CPAC-VGG uses labeled connections and \cnet uses \% labeled  images/class.}
	\label{tab: unlab_perform}
}
\end{table*}
\begin{table*}[h!]
{\small 
	\centering
		\begin{tabular}{|c||c|c||c|c||c|c|}
		\hline
Labeled Data   & \multicolumn{2}{c|}{FRGC} & \multicolumn{2}{c|}{CMU-PIE} & \multicolumn{2}{c|}{YouTube Faces} \\\cline{2-7}
 /Class (\%)&  NMI & ACC (\%) &  NMI & ACC (\%) &  NMI & ACC (\%)\\\hline
2  & 0.884 $\pm$ 0.61& 83.32 $\pm$ 8.80  &1.00 $\pm$ 0.00& 100 $\pm$ 0.00& 0.938 $\pm$ 0.016&  90.54 $\pm$ 2.92\\\hline 	
5  & 0.951 $\pm$ 0.010 &94.57  $\pm$  0.97& 1.00 $\pm$ 0.00&100 $\pm$   0.00& 0.989 $\pm$ 0.004 & 98.54 $\pm$ 0.61\\\hline
10 & 0.971 $\pm$ 0.0052&97.01 $\pm$ 0.46&1.00 $\pm$ 0.00 &100 $\pm$ 0.00&0.997 $\pm$ 0.002& 99.70 $\pm$ 0.25\\\hline
\end{tabular}
    \vspace{0.1cm}
	\caption{Performance of \cnet  with different percentage of labeled data on completely unseen face data samples}

	\label{tab: oose_results}
}
\end{table*}
\begin{table*}[h!]
{\small 
	\centering
		\begin{tabular}{|c||c|c||c|c|}
		\hline
Labeled Data   & \multicolumn{2}{c|}{USPS} & \multicolumn{2}{c|}{MNIST} \\\cline{2-5}
 /Class (\%)&  NMI & ACC (\%) &  NMI & ACC (\%)\\\hline
1 & 0.933 $\pm$ 0.006 &  96.33b$\pm$ 0.35 &0.939 $\pm$ 0.018 &98.13 $\pm$	0.24\\\hline 
2 &0.959 $\pm$ 0.007&98.02 $\pm$ 0.44 &0.958 $\pm$ 0.003&98.38 $\pm$ 0.18\\\hline 
5 & 0.965 $\pm$	0.009& 98.43 $\pm$ 0.46&0.963 $\pm$ 0.001 &98.65 $\pm$ 0.05\\\hline 
10 &0.971 $\pm$	0.007 &98.64 $\pm$ 0.39 &0.97 $\pm$	0.003 &98.96 $\pm$ 0.12\\\hline 
\end{tabular}
    \vspace{0.1cm}
	\caption{Performance of \cnet  with different percentage of labeled data on completely unseen digits data samples}
	\label{tab: oose_digits}
}
\end{table*}
\begin{figure*}[h!]
	\begin{center}
			\subfigure[YTF dataset]{\includegraphics[width =0.24\textwidth]{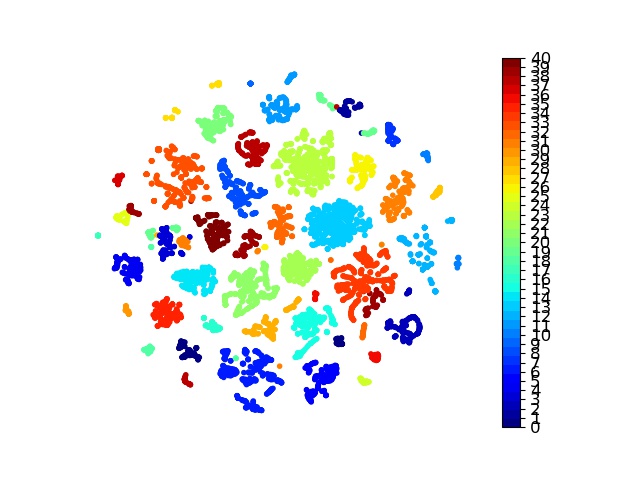}
        \includegraphics[width =0.24\textwidth]{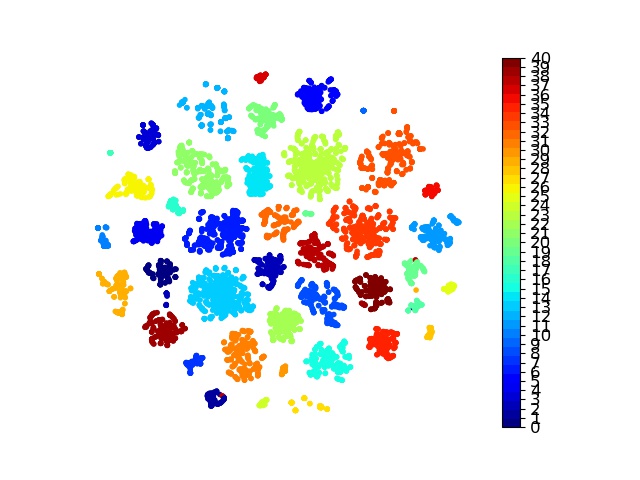}}\;
        \subfigure[MNIST dataset]{\includegraphics[width =0.24\textwidth]{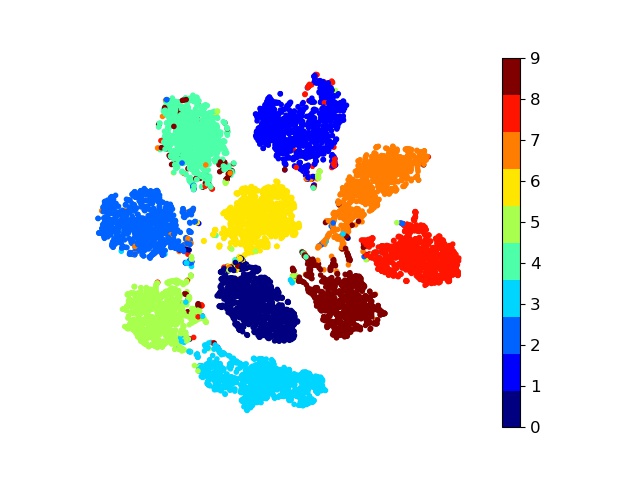}
        \includegraphics[width =0.24\textwidth]{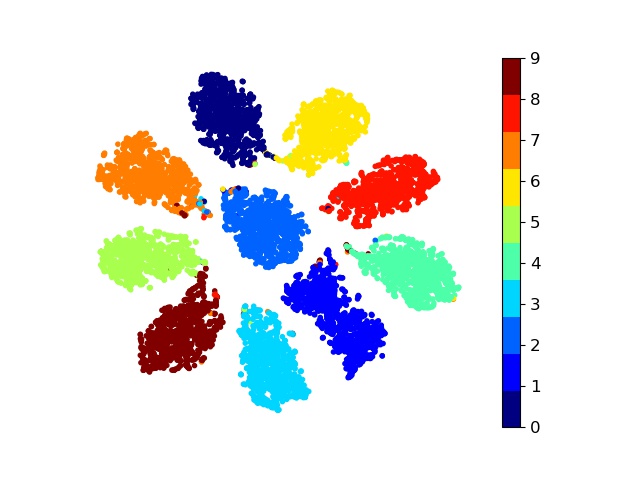}}
		\caption{T-sne plots for two different percent of labeled data (a) YTF (Left  : $2\%$ labeled data, Right: $5 \%$ labeled data) and  (b) MNIST (Left: $0.1\%$ labeled data, Right: $0.5 \%$ labeled data) }
		\label{fig:frgc_tsne}
		\end{center}
\end{figure*}

\noindent \textbf{Comparison with state of the art approaches}
\cnet is a semi-supervised approach that uses both labeled and unlabeled data. Since semi-supervised approaches are not explored for end-to-end clustering, we present comparisons with state of the art unsupervised clustering approaches as well as approaches that can be adapted to operate in the semi-supervised setting.



Fogel \emph{et al.} \cite{Fogel_pair2018}: This work proposes a non parametric approach that does not require to know the number of clusters a priori. The approach works in both unsupervised as well as semi-supervised setting. In order to bring supervision in the framework, while they define pairwise connections from unlabeled data, they either manually label or use ground truth to generate some data pairs as similar and dissimilar. A subset of these pairwise constraints drives the clustering algorithm. In Table \ref{tab: unlab_perform}, we present a comparative analysis of our results with the best reported results in \cite{Fogel_pair2018} with $5 k $ labeled connections.

Kira \emph{et al.} \cite{Kira_2015}: This approach utilizes labeled data to create similar and dissimilar constraints, similar to our approach. The paper also shows results when very few labeled samples are used to extract pairwise constraints from the data. Since they do not use any unlabeled data, in low training data setting, their performance is suboptimal. \cnet outperforms the best clustering accuracy reported for MNIST dataset by a margin of $15.71 \%$, $3.26 \%$ and $0.26 \%$ respectively for 6, 60 and 600 samples/class respectively.

The unsupervised approaches used for comparison are: JULE \cite{Batra_CVPR2016}, DEC \cite{Xie_ICML2016} and DEPICT \cite{Dizaji_ICCV2017}. 

\subsection{Network Details}
We used a common architecture for \cnet across all the datasets, and show that its clustering performance generalizes well. We used three convolution layers with 32, 64 and 128 filters respectively, followed by a fully connected layer of dimension 32. Each convolution layer is followed by InstanceNorm layer and Leaky ReLU activation, while the fully connected layers in both encoder and decoder use a tanh activation function. \added[id=SA,remark={what does `appropriately selected' mean here?}]{The stride and padding values are chosen appropriately for the dataset.} For each of the datasets, we pre-train the autoencoder end-to-end for 100 epochs using Adam optimizier, with a learning rate of $0.0001$ and $\beta =(0.9, 0.999)$ to minimize the reconstruction error. We use a dropout of $10\%$ at each layer, except the last layer of the decoder. We use this pre-trained network and fine tune it for clustering loss for 60 epochs, again with the Adam optimizer and a learning rate of $0.0001$. The value of balancing coefficient is determined by eq. \ref{eq: bal_val}.
\begin{align}
 \lambda = \begin{cases}
0 \quad &t< T_1\\
   \frac{t-T_1}{T_2-T_1} \quad  &T_1\leq t \le T_2\\
   1  \quad &T_2 \leq t
  \end{cases}
  \label{eq: bal_val}
\end{align}
Here, $t$ denotes the current epoch, $T_1 =5$, $T_2 = 40$ for all the experiments.
\subsection{Evaluation Metric}
We use two standard metric to evaluate the performance of proposed approach: NMI (normalized mutual information) that computes the normalized similarity measure between between true and predicted labels for the data and is defined as
\begin{align}
NMI(c,c') = \frac{I(c;c')}{max(H(c),H(c'))}
\end{align}
Here, $c$ and $c'$ represents true and predicted class label, $I(c,c')$ is the mutual information $c$ and $c'$ and $H(\cdot)$ denotes the entropy.
We also report the classification accuracy (ACC) that gives the percentage of correctly labeled samples. For our approach, the predicted label is the cluster with minimum distance from the given sample point.

\subsection{Clustering and Classification Results}
We compare the performance of \cnet with other clustering approaches on all the datasets. For our approach, we show both the average as well as the best achieved over five runs. For each dataset except MNIST, we first randomly split the data using stratified sampling with 90\% for training and hold-out 10\% for testing. We further generate five random splits from training data using stratified sampling into labeled and unlabeled data.  For other approaches, we quote the results reported in the original reference. Table \ref{tab: unlab_perform} summarizes both the classification as well as the clustering performance comparisons. The results show that \emph{ClusterNet} outperforms state-of-the art approaches by using small amounts of labeled data ($0.5$ or $2 \%$). The high NMI scores for \cnet can also be corroborated by the clusters formed in the t-sne based latent space visualization shown in Fig. \ref{fig:frgc_tsne}  for the YouTube Face and MNIST datasets. The visible structure in the figures is typical and similar visualizations are obtained for other datasets as well. 

We also show the effect of training the pre-trained autoencoder with the clustering loss in Fig. \ref{fig:before_after} and the corresponding NMI results in Table \ref{tab: kmeans}.  As before, the t-sne plots shown on the FRGC dataset is a typical indication of a clusterable latent space. 
\begin{figure}[h]
	\centering
		\subfigure[CMU-PIE]{\includegraphics[width =30mm]{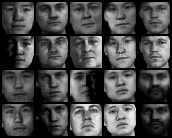}}
        \;
        \subfigure[FRGC]{\includegraphics[width =30mm]{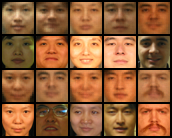}}\;
		\caption{Randomly selected 10 decoded cluster center images with respect to  cluster center representations in the latent space. Cluster Center Images: First and Third row, Sample Images: Second and Fourth row}
		\label{fig:cmu_centers}	
\end{figure}
\begin{figure}[h]
	\centering
		{\includegraphics[width =40mm,scale=0.8]{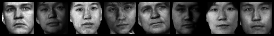} 
        \includegraphics[width =40mm,scale=0.8]{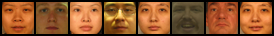}}
        {\includegraphics[width =40mm,scale=0.8]{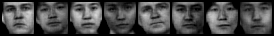} 
        \includegraphics[width =40mm,scale=0.8]{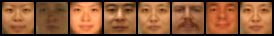}}
		\caption{CMU-PIE and FRGC original images in top left and top right respectively with corresponding reconstructed images in bottom left and bottom right}
		\label{fig:pie_data}
\end{figure}
\begin{table}[h!]
{\small 
	\centering
		\begin{tabular}{|c|c|c|c|}
		\hline
        Datasets & Autoencoder & \multicolumn{2}{c|}{\cnet} \\\cline{2-4}
        & All data & Train & Test\\\hline
        CMU-PIE & 0.788 $\pm$ 0.006 &1 $\pm$ 0.0  & 1 $\pm$ 0.0 \\
        FRGC &0.485 $\pm$ 0.012 &0.945 $\pm$ 0.002 & 0.951 $\pm$ 0.003\\
        YTF & 0.761 $\pm$ 0.003&0.949 $\pm$ 0.0 &0.953 $\pm$ 0.0  \\
        USPS &0.491 $\pm$ 0.001 &0.931 $\pm$ 0.0  &0.961 $\pm$ 0.0  \\
        MNIST &0.542 $\pm$ 0.0 & 0.942 $\pm$ 0.0 & 0.954 $\pm$ 0.0 \\\hline
        \end{tabular}
        \vspace{0.1cm}
	\caption{Comparison of Kmeans clustering performance (NMI) of autoencoder embeddings with \cnet embeddings }
	\label{tab: kmeans}
}
\end{table}

\begin{figure*}[h!]
	\centering
    \subfigure[]{\includegraphics[width=0.3\textwidth]{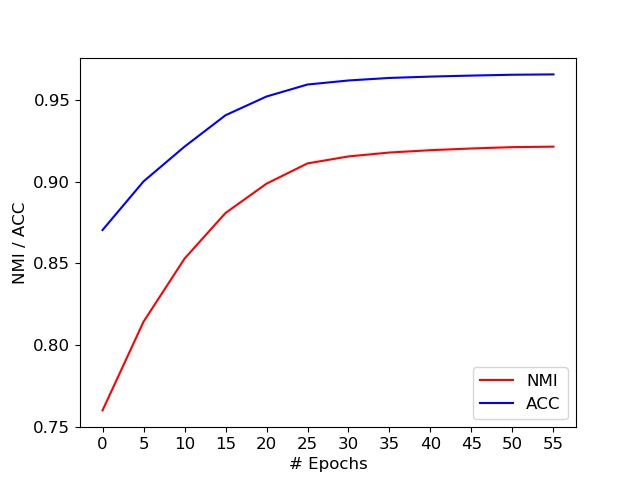}\label{fig:acc_nmi_usps}}
    \subfigure[]{\includegraphics[width =0.3\textwidth]{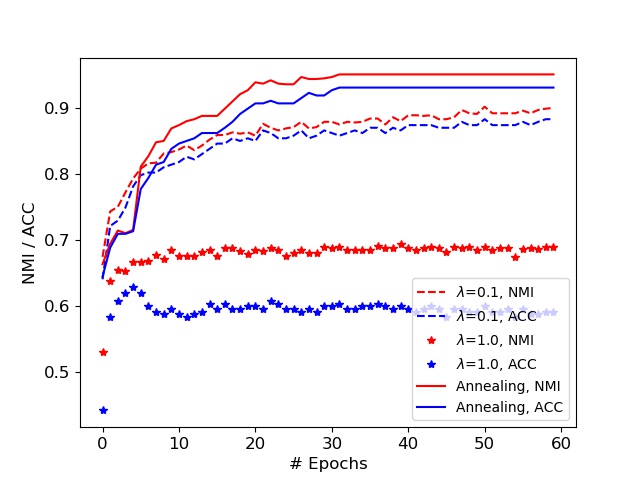}\label{fig:lamda_frgc}}
  \subfigure[]{\includegraphics[width =0.3\textwidth]{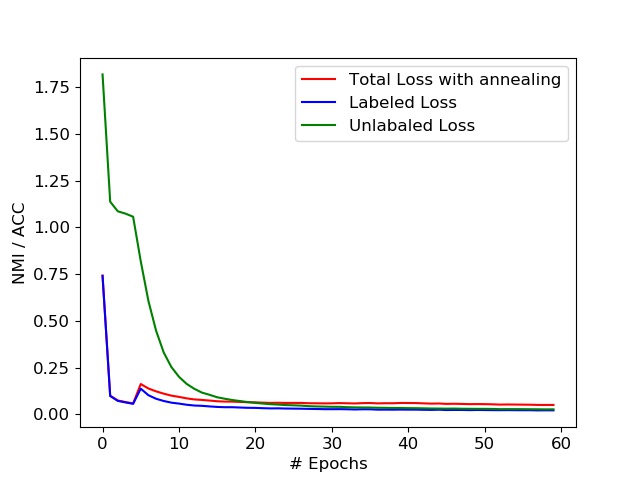}\label{fig:ytf_loss}}
\caption{(a) Consistency in NMI and ACC over epochs for unlabeled USPS digits training data, (b) Effect of $\lambda$ (balancing coefficient) on NMI and ACC on FRGC dataset (c) Loss function plot for YTF dataset with $2 \%$ labeled data}	 	
\end{figure*}

\subsection{Analysis}
\noindent \textbf{\textit{Cluster Purity during Training}}. As shown in Algo. \ref{algo:clusternet}, at training time, we make use of predicted labels for the unlabeled data to compute the pairwise KL-divergence. It is crucial for most predicted labels to be relatively consistent with the true class labels, or the training may diverge. In Fig. \ref{fig:acc_nmi_usps}, we plot the test accuracy and NMI over the training epochs for the USPS dataset as a typical example. We point out that the annealing strategy \eqref{eq: bal_val} applied to the balancing coefficient $\lambda$ is important for stabilizing the clustering. 

\noindent \textit{\textbf{Effect of Balancing Coefficient}}. In Fig. \ref{fig:lamda_frgc}, we show the effect of $\lambda$ on the test accuracy and NMI values on the FRGC dataset. Since the labeled data are typically much smaller than the unlabeled data, it is important to balance the loss contributed by unlabeled data for achieving good representations and clustering performance. Too small a value of $\lambda$ underutilizes the unlabeled data, while too large a value may lead the training loss to be too high. The performance is significantly better when an annealing strategy is used to adapt $\lambda$ throughout the training process. Further, we also observe a smooth convergence as shown in the loss function plot for Youtube Face dataset in the Fig. \ref{fig:ytf_loss}.

\noindent \textbf{\textit{Cluster Centers}}
Since our approach uses labeled data for initializing the cluster centers and also update them using the corresponding labeled data samples over the epochs, each cluster center is representative of one of the true classes decided by the labeled samples. This one-to-one correspondence between clusters and classes provides the flexibility to use the cluster centers directly for classification of unseen data points.
We show the reconstructions of the learned cluster centers and a sample face image from the same class for two of the datasets in Fig. \ref{fig:cmu_centers}.
            


\noindent \textbf{\textit{Performance with Varying Label Data}}
We present our results for the face and digit datasets, with varying number of labeled samples in Tables \ref{tab: oose_results} and \ref{tab: oose_digits} respectively. The  number of data samples used for testing for each of the datasets is given in Table \ref{tab: datasets_oose}. We use a maximum of $10 \%$ labeled data for training the network, but our technique significantly improves the performance over unsupervised approaches for most datasets with as low as $2 \%$ labeled data. Further, our approach is more suitable for practical clustering tasks as opposed to existing semi-supervised approaches, as supervision is provided with notably fewer samples, and yet boosting the performance significantly. However, as we see in Table \ref{tab: oose_results}, there is a huge deviation in performance for FRGC dataset with $2 \%$ samples/class due to large class imbalance. Here, the lowest number of samples in a class is only $6$, which results in choosing only 2 samples from the class as labeled data.

\noindent \textbf{\textit{Image Reconstruction Results}}
\cnet  is an autoencoder based model, and we use the reconstruction loss as described in Sec. \ref{sec:obj_fn} for better generalization. We show some example images for the unseen test set of CMU-PIE and FRGC dataset in Figure \ref{fig:pie_data}. The reconstruction quality indicates that the learned latent space is not degenerate and the decoder can successfully generate the images from the embeddings.

\section{Conclusion}\label{sec:conc}
In this work, we introduced a convolutional autoencoder based semi-supervised clustering approach that works in an end-to-end fashion. The clustering and representation learning is driven by the few labeled samples as well as the unlabeled data and their predicted cluster-membership labels, as they evolve during the training process. The loss comprises a $k$-means style cluster loss and a pairwise KL-divergence term regularized by the autoencoder's reconstruction loss. Further, experimental results show that \cnet achieves better performance over state of the art unsupervised and semi-supervised approaches using only $2 \%$ of labeled data. We achieve stability in training by employing an annealing based strategy to adjust the balancing coefficient. 

In Sec. \ref{sec:expts}, using empirical analyses, we have demonstrated that with only a small amount of labeled data and a lot of unlabeled data, it is possible to achieve good clustering and classification performance. While the proposed approach does perform well in clustering, its classification accuracy is still lower than the state of the art semi-supervised \emph{classifier} presented in \cite{rasmus2015semi}. In order to close this gap, one possible future direction is to use probabilistic predictions of labels and perform clustering in a Bayesian framework. 

{\small
\bibliographystyle{ieee}
\bibliography{ClusterNet_bib}
}

\end{document}


\title{Supplementary Material}
\author{First Author \\
Institution1\\
{\tt\small firstauthor@i1.org}
\and
Second Author \\
Institution2\\
{\tt\small secondauthor@i2.org}
}

\maketitle
\ifwacvfinal\thispagestyle{empty}\fi

\section{Additional Results}
We